\documentclass[a4paper]{article}
\usepackage{hyperref}
\hypersetup{colorlinks=true, allcolors=blue}

\usepackage[english]{babel}
\usepackage[utf8x]{inputenc}
\usepackage[T1]{fontenc}

\usepackage[a4paper,top=3cm,bottom=2cm,left=3cm,right=3cm,marginparwidth=1.75cm]{geometry}
\usepackage{amsmath}
\usepackage{graphicx}
\usepackage{fancyhdr}
\usepackage[colorinlistoftodos]{todonotes}

\usepackage[frozencache,cachedir=.]{minted}

\date{October 31, 2018}

\title{Robotic Room Traversal using Optical Range Finding}
\author{(TR2018-991)\\\\Cole Smith\\Eric Lin\\Dennis Shasha}

\begin{document}

\maketitle
\newpage


\begin{abstract}
Consider the goal of visiting every part of a room that is not blocked by obstacles. Doing so efficiently requires both sensors and planning. Our findings suggest a method of inexpensive optical range finding for robotic room traversal. Our room traversal algorithm relies upon the approximate distance from the robot to the nearest obstacle in 360 degrees. We then choose the path with the furthest approximate distance. Since millimeter-precision is not required for our problem, we have opted to develop our own laser range finding solution, in lieu of using more common, but also expensive solutions like light detection and ranging (LIDAR). Rather, our solution uses a laser that casts a visible dot on the target and a common camera (an iPhone, for example). Based upon where in the camera frame the laser dot is detected, we may calculate an angle between our target and the laser aperture. Using this angle and the known distance between the camera eye and the laser aperture, we may solve all sides of a trigonometric model which provides the distance between the robot and the target.
\end{abstract}


\section{Problem Statement}

How can a robot make an efficient traversal of a room with the least amount of passes over the room, and how do we measure distances from obstacles such that the robot traverses all human-accessible areas within the room? 


\section{Related Work}

The complete traversal of robotics through different terrain is not a new problem, and there has been similar work done before. However, these approaches require advanced hardware. We instead propose a cost-efficient manner of room traversal using more common materials. 

In a paper from the IEEE 2000 International Conference on Intelligent Robots and Systems, C. Eberst et al. showed that a robot could successfully travel through doorways and avoid obstacles using a multiple-camera array. In addition to optical methods, Eberst et al. also utilizes ultra-sonic sensors and laser scanning for increased navigation reliability \cite{20000575611}. Our implementation differs from these solutions in that we minimized the variety and cost of required sensors. 

Another such approach was conducted by Jonathan Klippenstein and Hong Zhang from the University of Alberta, Canada. Klippenstein and Zhang performed research in feature extraction from visual simultaneous localization and mapping solutions (vSLAM) \cite{894673}. Similarly, Alpen et al. at the 8th IFAC Symposium on Intelligent Autonomous Vehicles in 2013 explored SLAM features for Unmanned Autonomous Vehicle (UAV) flight for indoor robotic traversal \cite{ALPEN2013268}, and Sergio García et al. from University of Alcala proposes a solution for aerial vSLAM in a single-camera approach \cite{7781977}. Our approach uses simpler image transformations and filtering, meaning that our methods are affected less by low computational power, and low camera resolution. 

Regardless of the sensory approach for the complete traversal of spaces by robots, Edlinger and Puttkamer at the University of Kaiserslautern propose a solution for an autonomous vehicle to build an internal, two-dimensional map of the traversal space with no prior knowledge about the traversal space itself. In addition to their traversal approach, Edlinger and Puttkamer also leverage optical range finding for navigation \cite{EDLINGER}. Our approach is only concerned with the traversal geometry of the robot, so the room geometry need not be stored for our algorithm. 

\newpage

\section{Materials}

\subsection{Hardware}
\begin{itemize}
\item ARM, x86-64 OS for Go programs
\item iPhone 6s Plus
\item Generic iPhone Suction Car Mount
\item Stepper Motor (Any step count, 12v)
\item iRobot Roomba model 600
\end{itemize}

\subsection{Software}
\begin{itemize}
\item Room Traversal Algorithm
\item EasyDriver board for Stepper Motor
\item GPIO Driver for Roomba
\end{itemize}
\subsection{Cost Analysis}

The system is designed to be as cost effective as possible. Our camera solution total cost was less than \$90. The cost, without camera or robotic platform, can be broken down as:

\begin{itemize}
\item Raspberry Pi Model 3B: \$35
\item Generic Stepper Motor: \$12
\item EasyDriver Stepper Motor Driver: \$14
\item Generic iPhone Suction Car Mount: \$8
\item Plastic housing: \$11
\item \bf Total Approximate Cost: \$80
\end{itemize}

An iPhone is not required to use the camera interface. Any camera can be used so long as it can export PNG files to our system. For example, the Raspberry Pi Camera retails for \$26 as of 2018. Comparable LIDAR solutions can cost > \$300\cite{huang}.


\section{Hardware Implementation}

\begin{figure}[ht]
\includegraphics[height=3in, angle=-90]{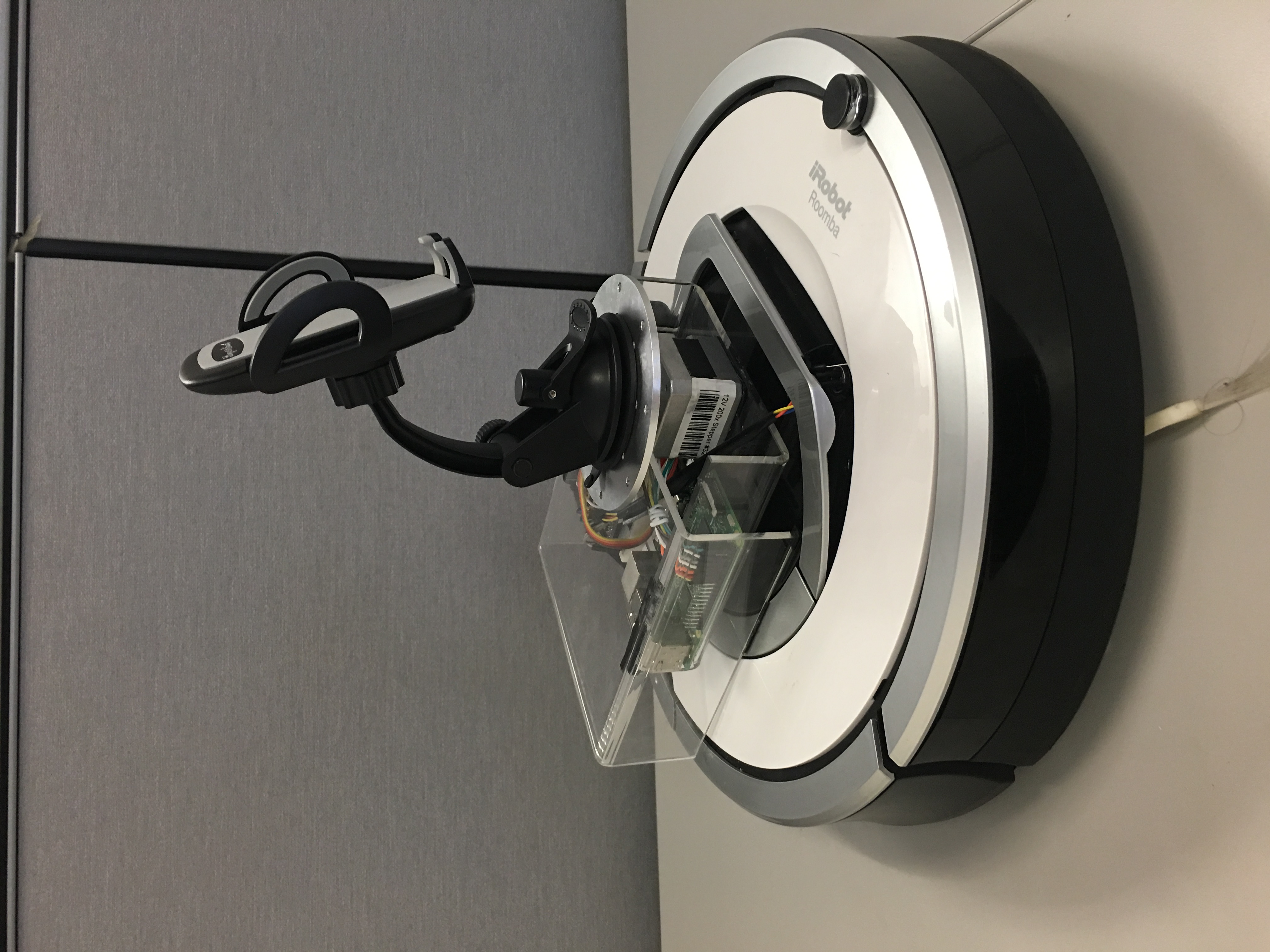}
\centering
\caption{Robotic platform using Roomba 600 model}
\label{fig:robot}
\end{figure}

\subsection{Robotic Testing Platform}

For our robotic platform, we are using an iRobot Roomba Model 600. Direct control is assumed over the Roomba using the provided serial port at the top of the unit. Our system implements a module that sends debug commands over the Raspberry Pi GPIO pins as serial output.

The Roomba is set to manual-drive mode using a specific serial command, and then subsequent move commands are sent to the Roomba when required. Since the Roomba is always set to move at a constant speed, location can be measured by counting the encoder values of the Roomba's wheels and comparing the value to the total time that the wheels are turning. 

\subsection{Camera Testing Platform}

The flow of image processing is as follows:

\begin{enumerate}
\item Image is streamed from iPhone to program as PNG, using HTTP server and client programs. (CameraStreamer iPhone Application)
\item PNG is converted to 2D pixel array in HSV color space.
\item Array is passed through luminosity thresholding filter.
\item Array is passed through color thresholding filter.
\item Blob detector is run on array, and the centroid of the blobs are detected.
\item Ovals are rejected from the system, leaving only the laser dot centroid.
\item The offset from the vertical center of the camera plane and laser dot is modeled as an angle, and used to calculate distance from obstacle to robot. 
\end{enumerate}


\section{Laser Dot Detection}

\subsection{Data Flow and Filtering}

PNG files are streamed from the CameraStreamer iPhone application to our program and converted into a 2D matrix in HSV (Hue, Saturation, Value) color space. Pixel values are stored as a struct, and each Hue, Saturation, and Value variable are normalized to be in the range [0,1]. This matrix will undergo a series of filtering  steps to convert it into a binary image mask. Three filters are used to detect a laser dot within an image: Luminosity, Color, and Oval Rejection. Each filter defines a set of thresholds and target values for which to convert the HSV matrix into a boolean matrix.

\subsection{Luminosity Filtering}

The first pass through the HSV matrix filters away pixels of undesired luminosity. For our purposes, we select only the brightest pixels within the image, since those are likely to represent the laser dot in the image frame. The conversion function creates a mask where "true" values are defined for pixels above or equal to the threshold value, and "false" for pixels below the threshold value.

\begin{minted}{Go}
func (image ImageMatrix) ConvertToMonoImageMatrixFromValue(valueThreshold float64) 
*MonoImageMatrix
\end{minted}

The conversion function defines a method on the \mintinline{Go}{ImageMatrix} struct and takes its Value threshold as an argument, \mintinline{Go}{valueThreshold}. This float defines the minimum cutoff for the Value (of HSV) of a pixel, in normalized range of [0,1]. The function then returns a pointer to a \mintinline{Go}{MonoImageMatrix} struct. This struct masks pixels of insufficient luminosity.

\subsection{Color Filtering}

A second pass is made over the HSV matrix to filter Hue values within the HSV color space. A target hue and a hue threshold are defined in which pixels are masked if the absolute value of the difference between the hue and the target hue exceeds the hue threshold.

\begin{minted}{Go}
func (image ImageMatrix) ConvertToMonoImageMatrixFromHue(hueTarget, hueThreshold float64) 
*MonoImageMatrix
\end{minted}

The function defines another method on \mintinline{Go}{ImageMatrix}. A pointer to a \mintinline{Go}{MonoImageMatrix} is returned, as it was in the previous luminosity filtering, with the pixels that deviate further than \mintinline{Go}{hueThreshold} from \mintinline{Go}{hueTarget} masked, and represented visually as black.

The two masks, luminosity and color, are then combined into one \mintinline{Go}{MonoImageMatrix}. Rendered, a masked image of a laser dot will appear like the following:

\begin{figure}[ht]
\includegraphics[height=2in]{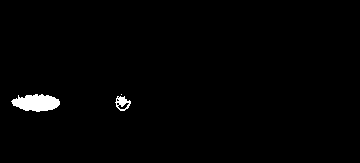}
\centering
\caption{A filtered image showing a green laser dot and reflection}
\label{fig:mask}
\end{figure}

\subsection{Blob Detection}

The masked matrix is then traversed using a 4-connected blob detection algorithm. Small artifacts can be rejected by defining a \mintinline{Go}{MIN_BLOB_SIZE} constant in number of pixels. The blobs are then returned as an array of pixel coordinate groups (X,Y) which are connected and "true" within the boolean image matrix. In figure ~\ref{fig:mask}, two groups of pixels will be returned.

\subsection{Oval Rejection}

The filtering passes will successfully mask out all light that does not conform to the luminosity and color profile defined as thresholds to the filtering methods. This will leave a boolean mask of the laser dot and any reflections of the laser dot in the image frame. Since these reflections will often appear less circular than the laser dot itself, we may reject the blobs that do not conform to a defined circular ratio.

\begin{minted}{Go}
// Given a series of connected coords, take the difference of
// min and max values for X and Y. The differences for X and Y
// are made a ratio as:
//		[ abs(minX) - abs(maxX) ] / [ abs(minY) - abs(maxY) ]
// or
//		[ abs(minY) - abs(maxY) ] / [ abs(minX) - abs(maxX) ]
// A ratio of 1.0 denotes a perfectly square bounding rectangle,
// (a circle blob). Anything less, denotes the oval ratio
func getCircleRatio(blob []*coord) float64
\end{minted}

In figure ~\ref{fig:mask}, the leftmost oval will be rejected from consideration as a laser dot, leaving only the rightmost blob. The centroid of this pixel coordinates group is then calculated. The pixel distance of the centroid to the vertical center of the camera plane is then used to determine the physical distance between the laser dot and the laser diode. 

\section{Range Finding Model}

\subsection{Calibration}

The range finding program must first be calibrated before it may be used. Two methods may be used: (1) One calibration step is used, with the camera rotated with every distance reading during normal operation, to determine the angle needed to match the laser dot to the vertical center of the camera plane. (2) Multiple calibrations steps are used, with the camera not rotating while taking distance readings during normal operation, to determine the rate at which the laser dot moves away from the vertical center of the camera plane.

Using method (1), the user will place the robot such that the laser diode is 1 unit of distance (1 meter, 1 foot, etc.) away from the laser dot projected on a clean surface. The camera will then rotate using the stepper motor until the laser dot converges to the vertical center of the camera plane. The angle of rotation found in this calibration step defines a triangle for which the side opposite to the hypotenuse is one unit of measure. Subsequent measurements during normal operation will be defined as a continuous proportion of this unit measure.

Using method (2), the user will originally place the robot as they would in method (1). More than one calibration steps are used, and the user will then place the robot at 2, 3, and 4 units of measure away from the laser dot. At each calibration step, the distance of the laser dot to the vertical center of the camera plane is recorded, and the rate of change in pixel distance is determined as the robot moves further away from the laser dot. This method of calibration has the limitation that the physical distance reading during normal operation is only as accurate as the approximated rate of change in pixel distance determined by this calibration method. For distances greater than the N units of measure conducted during calibration, the distance measured will be extrapolated from this approximated rate of change. For this reason, our implementation supports method (1) only. 

\begin{figure}[ht]
\includegraphics[height=4in]{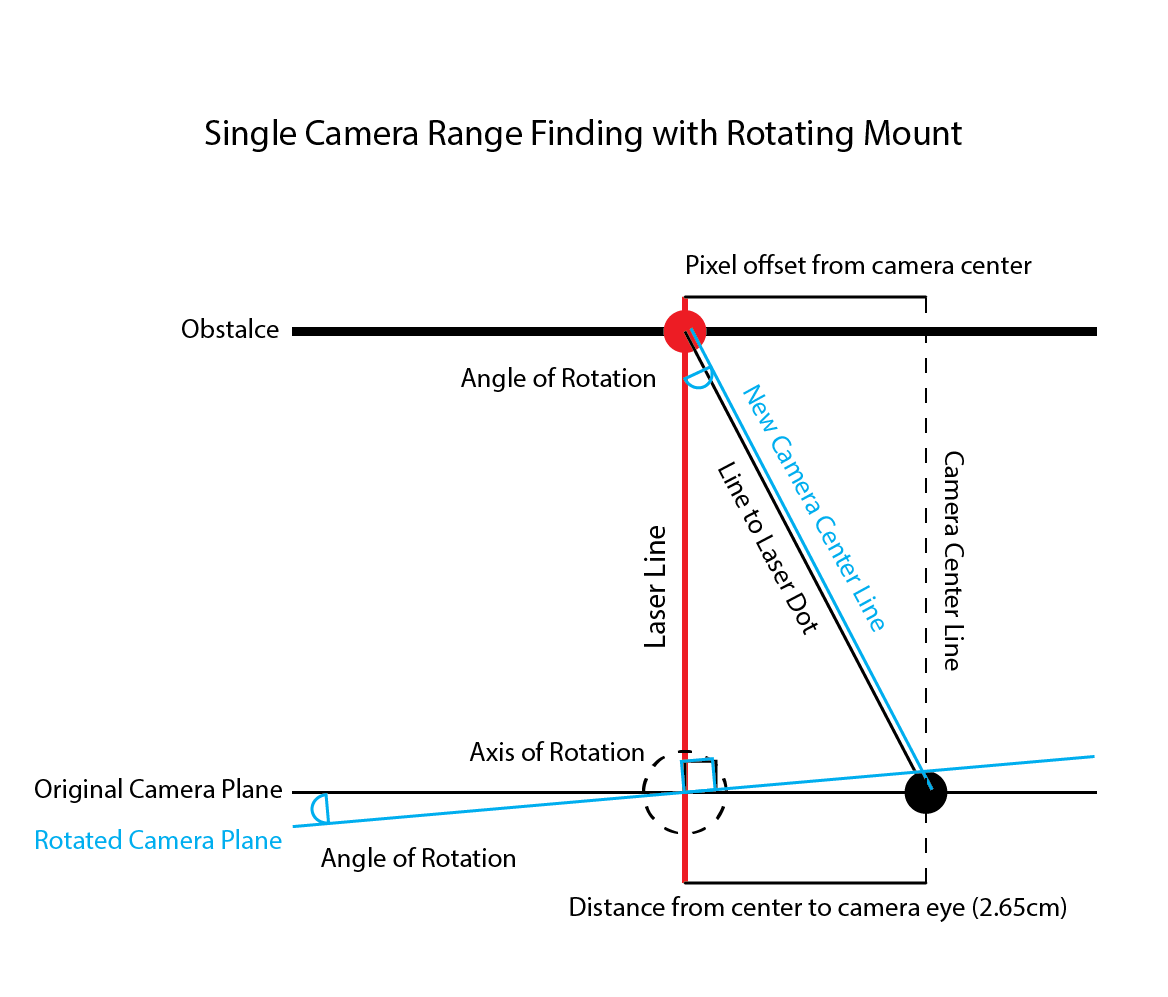}
\centering
\caption{Triangular Distance Model}
\label{fig:tri}
\end{figure}

\subsection{Range Calculation}

For each distance reading, the camera is rotated until the laser offset from the vertical center of the camera plane is minimized, or ideally zero. The angle of rotation required is recorded, and given to a simple function to solve the triangular model in Figure ~\ref{fig:tri} using the Sine Law. The camera rotation is performed using the stepper motor and iPhone mount seen in Figure ~\ref{fig:robot}

\begin{minted}{Go}
// Returns the distance from the laser diode to the target based upon the
// provided angle, at which the pixel offset was corrected by rotating the camera
// such that the laser dot was in the center plane of the camera.
func GetLaserDistance(angle float64, triangleBase float64) float64 {
	sineLawBase := (triangleBase / math.Sin(angle))
	sineOfAngleC := math.Sin(90 - angle)

	return sineLawBase * sineOfAngleC
}
\end{minted}

\subparagraph{Limitations}

The constraints of our range finding model have theoretical limits based upon camera resolution and stepper motor resolution (how many discrete "steps" are available in 360 continuous degrees). As camera resolution increases, the model can gauge distance further as the increased pixel count allows for greater room between the laser dot and vertical camera plane, such that the vanishing point (where the laser dot and vertical camera place will naturally converge as distance increases) will be further from the laser diode. While greater resolution will increase computation time polynomially, more possible steps within the stepper motor allow for a more accurate angle when rotating the vertical camera place to the laser dot.

\section{Room Traversal Method}

\begin{figure}[H]
\includegraphics[height=4in]{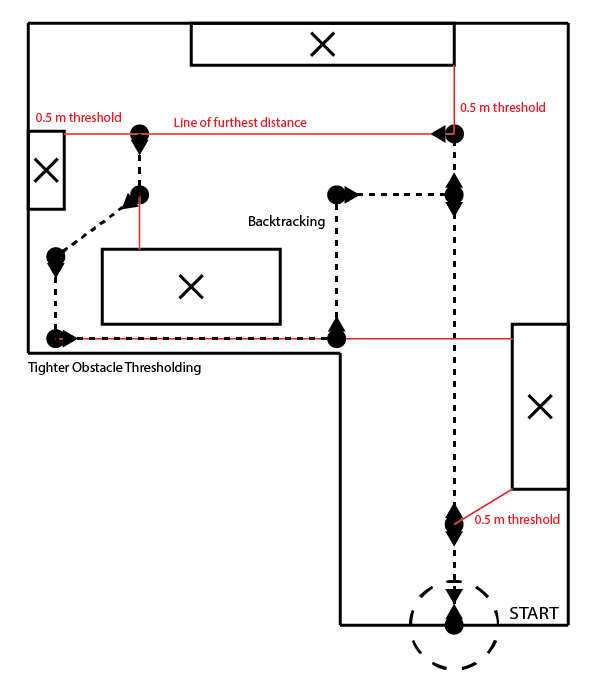}
\centering
\caption{Room Traversal Model with Range Finding}
\end{figure}

Based on the above image, the user will pick a starting point in a given room. The robotic vehicle will then scan the surrounding area in 60 degree increments for the direction it can travel the furthest, giving 6 possible directions of travel. Once a direction has been determined, it will start its navigation towards that direction, keeping a predetermined amount of threshold between the vehicle and other potential obstacles.

The vehicle continues down the direction until the threshold eventually stops the vehicle from traveling in that direction, and then it scans the room again for the furthest direction to travel without backtracking.

The vehicle will eventually reach a point where it cannot move forward without backtracking, and once that point is reached, it will first decrease the obstacle threshold and determine whether it allows the vehicle to move in additional spaces it has not been to before. An algorithm can be defined as follows:

\begin{enumerate}
\item START at doorway or accessible entrance
\item Take 6 distance readings in 360 degrees and begin traversing the path of maximum distance
\item Stop when distance to obstacle in the forward travel direction is less than threshold distance
\item Take 6 distance readings in 360 degrees
\item If all distance readings are below threshold, temporarily lower threshold to maximum distance of previous reading
\item If threshold has been lowered to less than the width of the robot (the robot can no longer traverse into a space), backtrack out of space by following previous line of travel, goto 4. 
\item Else if forward movement crossed a line of previous traversal (cycle detected), stop, find path to starting position using previously traversed paths, follow path, END.
\item Else, goto 2

\end{enumerate}

Once all available space has been traversed, it will then return to the starting position through the nearest path it can find to return.


\section{Results}

We compared our approach to a naive-bounce approach, in which the robot will make 90 degree turns when bumping into an obstacle. The traversal concludes when the robot reaches its starting position. 

We found our approach to offer improved traversal since it prevents the robot from entering infinite bounce-loops, or ending its traversal early, as seen in the figure below.

The below results compare our algorithm to the naive bounce approach. The grey blocks represent obstacles. Yellow lines denote the robots traversal path, which begins in the lower-right corner. For our algorithm's approach, the initial obstacle threshold was set to 20 units. The 

\begin{figure}[ht]
    \centering
    \includegraphics{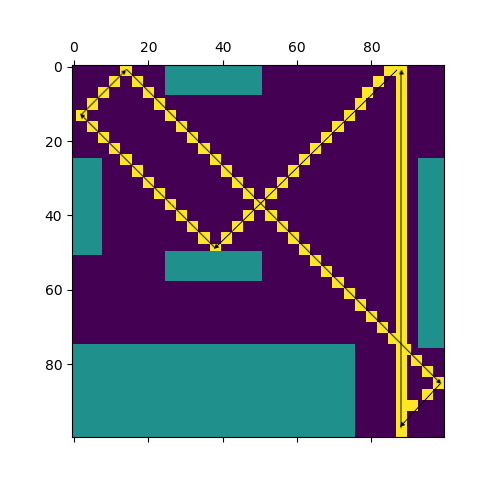}
    \caption{Robot Traversal using Naive Bounce (90 Degrees)}
    \label{fig:bounce}
\end{figure}

\begin{figure}[ht]
    \centering
    \includegraphics{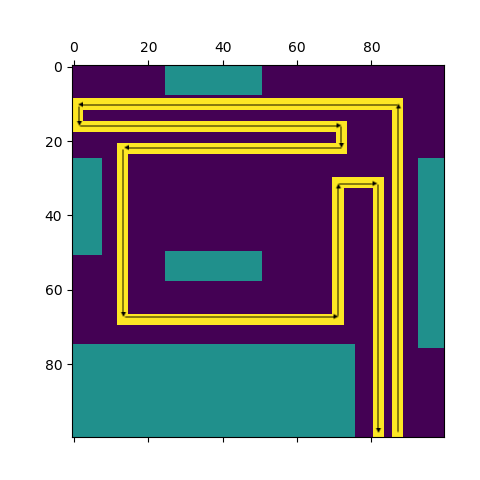}
    \caption{Robot Traversal using our algorithm (threshold: 20 units)}
    \label{fig:my_label}
\end{figure}


\section{Limitation: No Outdoor Robotic Traversal}

Our current robotic platform will not perform well in rough terrain outside of an office or home setting. Our system assumes flat ground with distance determined in 2 dimensions around the robot. As the robot encounters rougher terrain, a pitch in the Y dimension will be introduced. In order to properly handle outdoor situations, our model will have to be adjusted to incorporate distance calculation in 3 dimensions. This could be achieved by measuring the pixel distance between the laser dot and the horizontal center of the camera plane.

\section{Parallel Camera Support}

Our language choice (Go) naturally allows for more than one camera to operate in parallel. As a further step, more cameras and mounts can be added behind, or to the sides of the robot to decrease the need for the robot itself to rotate to gauge distance in 360 degrees. Currently, the laser itself is fixed, so the robot must rotate its entire assembly to point the laser at a different obstacle. Additionally, distance can be determined in 3 dimensions by adding cameras and lasers pointing upwards or at a pitched angle. 


\section{Conclusion}

The current mono-camera SLAM has been tested and shown to be successful in a flat indoor areas. Our system provides an autonomous and cost-effective solution to room traversal in stable environments. For further considerations, we would like to improve the ease of threshold tuning for the machine vision pipeline model, and expand the CameraStreamer iOS application to provide control over the entire system. In doing so, we would allow our system the first steps into less stable environments such as outdoor scenarios with high brightness. In extremely bright environments, our model will support the use of IR lasers and cameras. Additionally, the pitch of the camera in the Y dimension would need to be considered for non-flat terrain. Given these issues are addressed, this research provides future expansion into areas such as unmanned aerial vehicle navigation, since our long term considerations include abstracting our model to a more general method of environment traversal.

We provide a basic SLAM scaffold for any robotic vehicle using a single camera setup on our repository home page:

\url{https://github.com/NYU-Efficient-Room-Traversal}


\newpage
\section{Acknowledgements}

This research was supported in part by the funding of NYU College of Arts and Sciences Dean's Undergraduate Research Fund and NYU WIRELESS.

\bibliographystyle{plain}
\bibliography{refs}
\nocite{*}

\end{document}